\documentclass{article}
\usepackage{natbib}
\setcitestyle{numbers,square}

\usepackage[final]{paper_arxiv}

\usepackage[utf8]{inputenc} 
\usepackage[T1]{fontenc}    
\usepackage{hyperref}       
\usepackage{url}            
\usepackage{booktabs}       
\usepackage{amsfonts}       
\usepackage{nicefrac}       
\usepackage{microtype}      
\usepackage{xcolor}         
\usepackage{graphics}
\usepackage{graphicx}
\usepackage{booktabs} 
\usepackage{wrapfig}
\usepackage{enumitem}
\usepackage{subcaption}
\usepackage{amsmath}
\title{GITA: Graph to Visual and Textual Integration \\
for Vision-Language Graph Reasoning}

\author{
Yanbin Wei\textsuperscript{\rm 1,2\thanks{Equal contribution.}},
Shuai Fu\textsuperscript{\rm 1,3\footnotemark[1]}, 
Weisen Jiang\textsuperscript{\rm 1,2\footnotemark[2]},
Zejian Zhang\textsuperscript{\rm 4},
Zhixiong Zeng\textsuperscript{\rm 4}, \\
\textbf{Qi Wu}\textsuperscript{\rm 3},
\textbf{James T. Kwok}\textsuperscript{\rm 2},
\textbf{Yu Zhang}\textsuperscript{\rm 1\thanks{Corresponding authors.}} \\ 
\textsuperscript{\rm 1}\small Department of Computer Science and Engineering, Southern University of Science and Technology \\
\textsuperscript{\rm 2}\small Department of Computer Science and Engineering, Hong Kong University of Science and Technology \\
\textsuperscript{\rm 3}\small Australia Institute for Machine Learning, University of Adelaide~~~
\textsuperscript{\rm 4}\small Tencent \\
\{\texttt{yanbin.ust, fus.jayce,  zejianzhang33, yu.zhang.ust, waysonkong\}@gmail.com} \\
\texttt{barretzeng@tencent.com},~
\texttt{qi.wu01@adelaide.edu.au},~
\texttt{jamesk@cse.ust.hk}\\
}

\begin{document}

\maketitle

\begin{abstract}
Large Language Models (LLMs) are increasingly used for various tasks with graph structures. Though LLMs can process graph information in the textual format, they overlook the rich vision modality, which is an intuitive way for humans to comprehend structural information and conduct general graph reasoning. The potential benefits and capabilities of representing graph structures as visual images (i.e., \textit{visual graph}) are still unexplored. To fill the gap, we innovatively propose an end-to-end framework, called \textbf{G}raph to v\textbf{I}sual and \textbf{T}extual Integr\textbf{A}tion (GITA), which incorporates visual graphs into general graph reasoning. Besides, we construct the \textbf{G}raph-based \textbf{V}ision-\textbf{L}anguage \textbf{Q}uestion \textbf{A}nswering (GVLQA) dataset from existing graph data, which is the first vision-language dataset for general graph reasoning. Extensive experiments on the GVLQA dataset and five real-world datasets show that GITA outperforms mainstream LLMs on general graph reasoning. Moreover, experimental results demonstrate the effectiveness of the layout augmentation on visual graphs and pretraining on the GVLQA dataset.
\end{abstract}

\section{Introduction} 
\label{sec:intro}


Graph reasoning tasks are pivotal in domains such as recommendation systems \citep{lightgcn,tui2}, social network analysis \citep{social1,huang2021knowledge}, and knowledge graph reasoning \citep{zhang2021cone,liu2022mask,llmkgc}. Various architectures have been developed, from shallow embedding methods \citep{bordes2013translating, socher2013reasoning} to advanced Graph Neural Networks (GNNs) \citep{kipf2016semi,xu2018powerful} and graph Transformers \citep{zhang2020graph,kreuzer2021rethinking,chen2022structure}. While these models excel in graph reasoning tasks, they often lack generalizability, flexibility, and user-friendliness. Achieving good performance with these models typically requires domain-specific tuning, which limits their abilities to generalize across different domains. Additionally, these models struggle to handle diverse tasks with the same architecture. Each task often requires a specialized design, including task-specific data processing and decoder, leading to limited flexibility. Lastly, unlike the Large Language Models (LLMs) that can engage in conversations with users, these models are less explainable and user-friendly.

In contrast, LLMs have shown great generalization capabilities across a wide variety of reasoning tasks \citep{wei2022chain,yu2023metamath,zhou2022least, jiang2024forward, jiang2023effective} by encapsulating task-specific demands within a cohesive and interpretable mechanism - text prompts, under a unified architecture with minimal domain-specific adjustments. 
These advantages have sparked investigations into the potential of LLMs for graph reasoning. 
Recent developments lend credence to the notion that LLMs can indeed interpret and manipulate graph-structured data through textual representations. For example, InstructGLM~\citep{ye2023natural}, GPT4Graph~\citep{guo2023gpt4graph}, and LLMtoGraph~\citep{liu2023evaluating} convert graphs into textual descriptions and then use these descriptions paired with queries to enable LLMs to generate accurate responses for graph reasoning tasks. Furthermore, the introduction of benchmarks such as GraphQA~\citep{fatemi2024talk} and NLGraph~\citep{canllm} is a testament to the growing interest in evaluating LLMs' effectiveness on graph reasoning tasks framed in natural languages. 

Despite the development of numerous methods and benchmarks for graph reasoning on LLMs, they often overlook the valuable vision modality, which is a natural means for humans to comprehend structural information and has demonstrated its success in various visual reasoning scenarios \cite{gqa,vcr,raven,clevr,shapes,nlvr2}. 
Consequently, the following questions arise: (1) \textit{Can incorporating visual information be beneficial for general graph reasoning scenarios?} 
(2) \textit{If so, how can we effectively integrate the vision modality into graph reasoning?} 
To the best of our knowledge, these questions remain unexplored.

To answer them, we first propose an end-to-end framework called \textbf{G}raph to v\textbf{I}sual and \textbf{T}extual Integr\textbf{A}tion (GITA)\footnote{Project Homepage: \url{v-graph.github.io}.}\footnote{Code Repository: \url{https://github.com/WEIYanbin1999/GITA/}.} that systematically integrates visual information into instruction-based graph reasoning, by rendering graph structures to customized visual images which are called \textit{visual graph}. 
Specifically, the GITA framework has four components: a \textit{graph visualizer} for generating visual graphs, a \textit{graph describer} for producing textual descriptions of the graph structure, a \textit{task-based questioner} that organizes the description and requirements of the current task into prompt instruction, and a \textit{Vision-Language Model} (VLM) to perform vision-language graph reasoning. In the proposed GITA framework, visual information can be incorporated into many tasks with explicit or implicit graph structures, without sacrificing its versatility, flexibility, or user-friendliness. Besides, since there is no dataset for vision-supported general graph reasoning capabilities, we construct the first vision-language dataset for general graph reasoning purposes called \textbf{G}raph-based \textbf{V}ision-\textbf{L}anguage \textbf{Q}uestion \textbf{A}nswering (GVLQA)\footnote{Dataset: \url{https://huggingface.co/collections/Yanbin99/}.} based on the proposed GITA framework. The GVLQA dataset consists of 526K instances covering seven representative graph reasoning tasks, aiming to thoroughly evaluate the structure-based graph reasoning abilities of VLMs and LLMs. Extensive experiments on the GVLQA dataset and five real-world datasets demonstrate the effectiveness of the proposed GITA model. Furthermore, we delve into the effects of visual graph augmentation strategies and find that layout augmentation can dramatically boost vision-based graph reasoning performance.

Our main contributions are summarized as follows.
\begin{itemize}
\item 
We introduce an end-to-end GITA framework, innovatively integrating vision modality to boost the graph reasoning abilities of language models.

\item 
We establish GVLQA, the first vision-language question-answering dataset for general graph reasoning purposes. It can be used to thoroughly evaluate the structure-based graph reasoning abilities of LLMs/VLMs and can also be used as pretraining data to boost the performance of downstream tasks.

\item Extensive experiments on benchmark datasets across various graph reasoning tasks demonstrate the effectiveness of the proposed GITA framework and  the benefits of layout augmentation on visual graphs.

\end{itemize}

\section{Related Work}
\label{sec:related-work}

\noindent{\bf Graph Reasoning.}
Graph reasoning \citep{battaglia2018relational, wu2020comprehensive} aims to answer questions based on graphs, which involves utilizing graph structures to guide the reasoning process to generate answers. Graph reasoning has a wide variety of applications in social network analysis \citep{newman2003structure, leskovec2008microscopic}, bioinformatics \citep{jeong2001lethality,gavin2006proteome}, chemistry \citep{gilmer2017neural}, physics \citep{battaglia2016interaction}, knowledge graph reasoning \citep{bordes2013translating}, and recommendation systems \citep{koren2009matrix, he2017neural}. 
Many 
graph reasoning
methods have been proposed. 
Early attempts \cite{bordes2013translating, socher2013reasoning} learn node and edge representations through shallow modules,
which may have only limited expressive power.
Graph Neural Networks (GNNs) such as  GCN \cite{kipf2016semi}, GAT \cite{velickovic2017graph}, GraphSAGE \cite{hamilton2017inductive}, MPNN \cite{gilmer2017neural}, and GIN \cite{xu2018powerful} use message-passing paradigm \cite{gilmer2017neural} to model graph dependencies and update node features. 
Transformer-based graph models \cite{zhang2020graph,kreuzer2021rethinking,chen2022structure} further propose to use self-attention to increase the expressiveness and long-range dependency. However, as discussed in Sec~\ref{sec:intro}, these models may exhibit limited generalizability, flexibility, and user-friendliness.

\noindent{\bf LLMs on Graph Reasoning.}
There have been many attempts to use LLMs in graph reasoning. 
Depending on how they align the input spaces of graphs and LLMs,
we categorize them into two types: Graph-to-text and Graph-to-token.
Graph-to-text methods transform a graph into textual descriptions,
which are concatenated with the instructions and fed to the LLM for querying. For example,
InstructGLM~\citep{ye2023natural} 
uses natural language to describe the graph 
and proposes instruction prompts to fine-tune the LLM. He et.al \cite{he2024harnessing} applies LLMs to explain graphs for training GNNs,
while Chen et.al \cite{chen2023exploring} treat LLMs as enhancers to exploit text attributes or as predictors for node classification on text-attributed graphs.
GPT4Graph \citep{guo2023gpt4graph} and LLMtoGraph \citep{liu2023evaluating} convert graphs into specific code or natural language formats by the powerful ChatGPT \citep{chatgpt,gpt4}.
On the other hand, Graph-to-token methods include GraphGPT~\citep{tang2023graphgpt}, GraphToken~\citep{perozzi2024let} and LLaGA~\citep{chen2024llaga}. For these methods, the graph is represented as a specially designed token sequence, which is projected or merged into the LLM's token space for
text-based reasoning. However, none of the aforementioned methods represent the graph structure information as images, highlighting the uniqueness of 
the proposed GITA framework and GVLQA dataset.

\noindent{\bf Large Vision-Language Models.} 
Large VLMs have significantly expanded the cognitive abilities of LLMs by integrating the vision modality to address vision-language tasks. 
Many methods have been proposed.
Some early explorations 
like Flamingo \cite{flamingo}, CLIP \cite{clip}, and BLIP-2 \cite{blip2} 
use
a visual encoder for processing images and align the visual and textual embeddings.
Subsequent models like  LLaVA \cite{llava} and MiniGPT-4 \cite{minigpt4} 
combine visual and textual inputs in a single LLM for solving multimodal tasks.
InstructBlip \cite{instructblip} proposes an instruction-aware query transformer and trains a vision-language model by instruction tuning. 
However, despite progress in a wide range of vision-language tasks \citep{visual_tasks_survey, nemesis}, using visual information in graph reasoning remains overlooked. We take the first step in this field, pushing the boundaries of VLMs in graph reasoning.

\section{GITA: Graph to Visual and Textual Integration}
\label{generate}

\subsection{Preliminary}
\noindent{\bf Graph Reasoning.}
In traditional graph reasoning settings, models typically rely on two main inputs: 
(i) the graph structure \(G = \{C, E\}\), where \(C\) and \(E\) are the set of vertices and edges, respectively; 
(ii) the task requirement \(T\), encompassing specific operations or questions pertaining to the graph. 
Based on the information provided in \(G\) and a specific task requirement \(T\), models are expected to output  
a reasonable answer \(A\). 
On the other hand, in the context of instruction-based graph reasoning methods, it is necessary to convert these inputs into textual form. This transformation facilities graph reasoning within natural language, allowing for improved interpretation and harnessing the formidable reasoning capabilities of large language models.



\subsection{Architecture}
\label{architecture}
\begin{figure*}[!t]
\centering
    \includegraphics[width=0.98\textwidth]{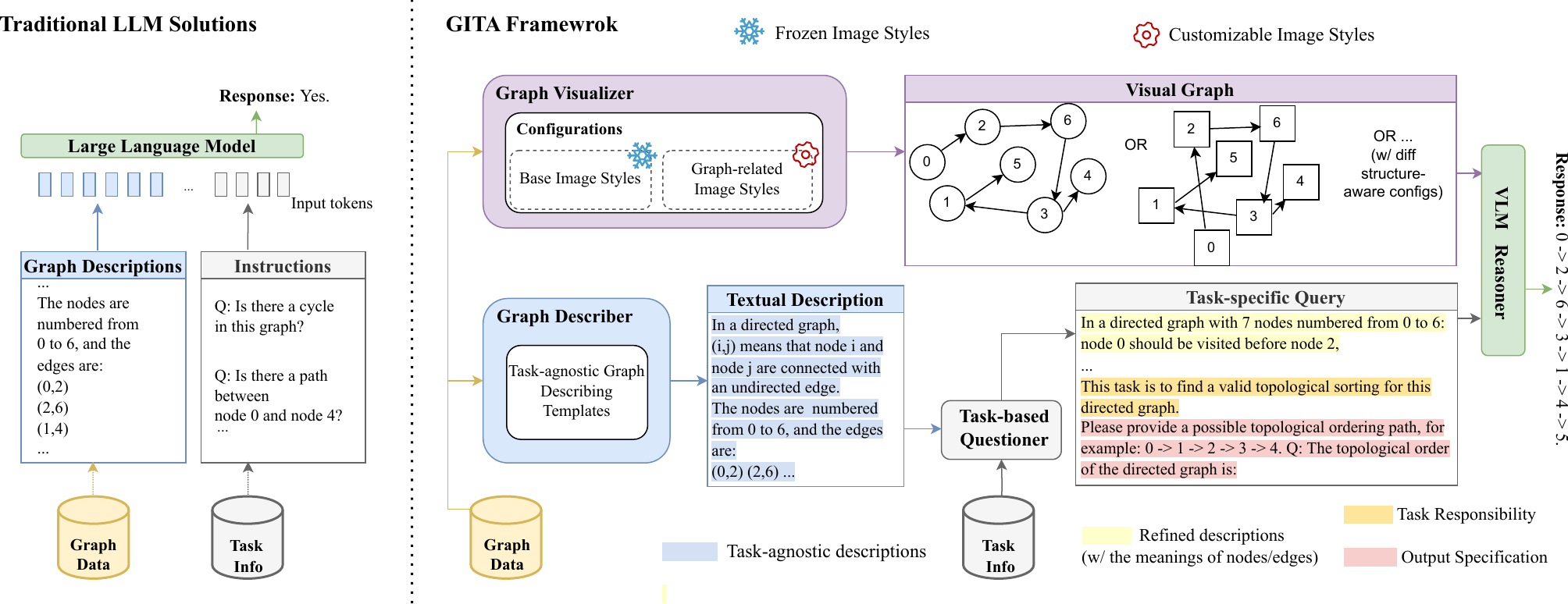}
        \vskip -.1in
    \caption{The architecture of the GITA framework with comparison to existing LLM solution.} 
    \label{fig:overview-method}
\end{figure*}

\noindent{\bf Overview.}
Different from the above graph reasoning methods, we propose a \textbf{G}raph to \textbf{I}mage-\textbf{T}xt \textbf{A}ssistant (GITA), which is the first attempt to perform graph reasoning in a vision-text-based manner. 
GITA comprises four pivotal components: a task-agnostic graph visualizer $V$, a graph describer $D$, a task-specific questioner $Q$, and a VLM reasoner $R_\phi$, as illustrated in Figure \ref{fig:overview-method}.
Firstly, $V$ and $G$ are designed to produce visual depictions (i.e., \textit{visual graphs}) and textual descriptions of the graph structure inputs, respectively. 
Then, given the task requirement $T$ and the textual description produced by $D$, $Q$ is designed to form a task-specific query. 
Finally, $R_\phi$ receives 
the visual input $I_G$ from $V$ based on 
the visual graph and the textual input $Q_G^T$ from $Q$, then 
generates answers $A$ in natural language.
In the following, we introduce the four components in detail.

\noindent{\bf Graph Visualizer.}
The role of the graph visualizer is to generate visual graphs from structural graphs. 
The image representation of a structural graph is not unique, as there can be variations in many aspects, such as backdrop colors, layouts, and node shapes. These variations may enhance the robustness of models through effective training but simultaneously increase the learning difficulty for models. Therefore, balancing \textit{consistency} and \textit{variety} is necessary during the graph visualization process. This trade-off is reflected in our design of graph visualizer, by maintaining consistency in basic image styles common to general images (i.e., size, resolution, backdrop) and only introducing customizable variations in four graph-related image styles unique to visual graphs (i.e., layout, node shapes, node outline styles, and edge thickness).
Graph visualization in $V$ can be formulated by the following equation:
\begin{equation}
\label{formulation:visual}
    I_G = V(G, \Gamma, \Delta),
\end{equation}
where \(I_G\) denotes the visual graph derived from graph \(G\), while $\Gamma$ and $\Delta$ are the fixed basic image styles and customizable graph-related image styles, respectively.

Visualizing the entire graph can be challenging when the number of nodes or edges is very large, affecting the clarity of the images. To address this, our graph visualizer adopts the standard strategy of \(k\)-hop subgraph sampling.
Specifically, \(k\)-hop subgraph sampling for a node \(u\) in the set of vertices \(C\) involves selecting a subgraph \(G_u = \{C_u \subseteq N_k(u), E_u \subseteq E\}\), where \(N_k(u)\) includes nodes within \(k\) steps from \(u\) and each edge \((i, j)\) in \(E_u\) connects nodes within \(C_u\). 
To generate the visual graph of the \( k \)-hop subgraph \( G_u \) centered on \( u \), the nodes within \( G_u \) are relabeled from 0 to \( |C_u| - 1 \) to facilitate the generalization of visual graphs. 
Subsequently, this relabeled subgraph \( G_u \) is fed to the graph visualizer to generate its visual graph $I_{G_u}$ by Eq. \eqref{formulation:visual}.

In practice, the graph visualizer can be implemented by a variety of graphic visualization tools, such as Graphviz \citep{graphviz}, Matplotlib \citep{matplotlib}, and NetworkX \citep{networkx}. Among them, Graphviz can automatically design the layouts of visual graphs, and is especially suitable for building large-scale datasets. Matplotlib is excellent for customizable plots with fine-grained control, and NetworkX excels in complex network analysis.
We have implemented various graph visualizers using modular, plug-in architecture in GITA. Specific examples of the visual graphs generated with these tools can be found in Appendix \ref{appendix: visualizer implement}.

\noindent{\bf Graph Describer.}
The graph describer $D$ is tasked with generating task-agnostic textual descriptions of a given graph $G$. To ensure clarity and fidelity of these descriptions, we meticulously craft a curated set of \textit{graph-describing templates}. The graph description templates outlined in Appendix \ref{appendix: describer template} are designed to cover a broad spectrum of scenarios, accommodating various graph configurations including directed or undirected graphs and those with or without node or edge weights. To generate the description for a given graph, the graph describer initially selects an appropriate template based on the graph's characteristics, such as its directionality and whether it includes node attributes or edge weights. 
Subsequently, this template is used by replacing placeholders with actual data, such as the number of nodes, the number of edges, and the endpoints of each edge, to craft detailed descriptions tailored to the specific graph in question.
The process for $D$ to generate textual descriptions can be formulated as follows:
\begin{equation}
    D_G = D(G, P),
\end{equation}
where \(D_G\) denotes the textual description generated by graph describer, and $P$ is the graph-describing template of the graph $G$.

By introducing these unified and structured graph-describing templates, the graph describer is empowered to generate coherent and informative descriptions that focus on the inherent characteristics of the graph itself, independent of specific task requirements.

\noindent{\bf Questioner.}
The questioner $Q$
is tailored to capture the intricate requirements of specific tasks and reflect them in its output task-specific query. 
In detail, $Q$ receives the task-agnostic textual descriptions from the graph describer and refines them to align with the task context by elucidating the concrete meanings of nodes and edges. 
These refined descriptions are then enriched with task responsibilities and input/output specifications to form task-specific queries. 
The formulation of the questioner to generate the task-specific queries can be represented as follows:
\begin{equation}
    Q_G^T = Q(T, D_G),
\end{equation}
where $Q_G^T$ represents the task-specific query generated by the questioner with given the task requirement $T$ and the textual description $D_G$.
The construction of task-specific queries can be approached in two main ways: manual template-based construction and bootstrapping LLM agents. 
Manual template-based construction enriches $D_G$ with task-specific manual templates, which is preferred for tasks with precise requirements, such as the Traveling Salesman Problem (TSP) \citep{tsp}, where accuracy is critical and the task definitions are well-understood. This is because it can ensure clarity and reduce the risk of errors due to its meticulous attention to details. 
On the other hand, bootstrapping LLM agents for automated synthesis is more economical and suitable for dynamic or bespoke tasks, such as robotic planning or complex gaming scenarios, as it can take advantage of the speed and adaptability of LLM agents to interpret context and generate appropriate queries, minimizing manual effort and enhancing responsiveness to changing conditions. 
Both methods are illustrated with examples in Appendix \ref{appendix: Questioner}, showcasing their applications and benefits in different scenarios.


\noindent{\bf VLM Reasoner.}
The VLM reasoner $R_\phi$ performs final graph reasoning with visual inputs $I_G$ from $V$ and textual inputs $Q_G^T$ from $Q$, and outputs responses in natural language. This reasoning process can be represented as the following:
\begin{equation}
    A = R(I_G, Q_G^T), 
\end{equation}
where $A$ is the answer generated by the vision-language model $R$.
In this work, we adopt GPT-4V and LLaVA-7B/13B as VLM reasoners. These models are regarded as representatives in the realm of closed-source and open-source VLMs, respectively. 


In summary, GITA systematically incorporates the vision modality into instruction-based graph reasoning. In Appendix \ref{appendix:characteristics}, we discuss the characteristics of GITA, in aspects of generalizability, flexibility and user-friendliness.

\subsection{Visual Graph Augmentation}
\label{visual aug}
Visual graphs generated for the same graph $G$ can be considered as an unique data augmentation technique. Building on the four graph-related image styles introduced in the graph visualizer part of Sec \ref{architecture}, we propose the following augmentation strategies: \textbf{layout augmentation}, \textbf{node shape augmentation}, \textbf{node outline style augmentation}, and \textbf{edge thickness augmentation}. 
Specifically, layout augmentation involves altering the layout styles while keeping all the other settings constant. Similarly, by changing only the respective attributes, we can implement node shape augmentation, node outline style augmentation, and edge thickness augmentation. These four proposed augmentation strategies facilitate studies on the importance of each in enhancing the graph reasoning abilities of VLM reasoners.

\subsection{Training}
Given a visual graph $I_G$ and a text-specific query $Q_G^T$, along with the target answer $A_t$, the VLM reasoner of GITA is trained to generate answers $A$. Specifically, $I_G$ is input into the vision encoder of the VLM reasoner, resulting in a set of visual features $F$. If there is a dimension difference between $F_v$ and the pretrained word embeddings, these $F$ will be aligned with the pretrained word embedding space of the text decoder by a vision-to-text projector. Finally, the aligned visual features $F_{aligned}$ and $Q_G^T$ are concatenated as input sequences of the text decoder.

Formally, given $I_G$, $Q_G^T$, and $A_t$, the VLM reasoner is trained by minimizing the following negative log-likelihood:
\begin{equation}
    \mathcal{L}_{\phi} = -\sum_{i=1}^{\lvert A \rvert} \log \ p_\phi (A_i \mid F_{aligned}, Q_G^T, A_{<i}),
\end{equation}
where $\phi$ is the trainable parameter and $A_i$ denotes the prediction token at the $i$-th position. Besides, $A_{<i}$ represents the first $i-1$ predicted tokens. During the inference process, GITA is capable of accepting structure graphs as inputs and performing graph reasoning in an end-to-end manner.


\section{GVLQA Dataset}
\label{GVLQA}
In this section, we introduce the GVLQA dataset to fill the absence of a vision-language-based general graph reasoning dataset. It is designed to:  1) evaluate the graph reasoning capabilities of VLMs or LLMs; 
2) help models acquire fundamental graph comprehension and reasoning abilities as a pretraining dataset.

\subsection{Construction}
The GVLQA dataset is created by utilizing the graph visualizerthe graph describer, and questioner in GITA to generate vision-language-based question-answer pairs for graph reasoning on an open-source graph dataset.
Specifically, we first extract both the original graph structures and the ground-truth outputs from the NLGraph-full dataset~\citep{canllm}. Then the graph visualizer (detailed in Sec \ref{architecture}) and the graph describer (outlined in Sec \ref{architecture}) are used to generate visual graphs and textual descriptions for these original graph structures, respectively. 
Afterwards, the questioner (described in Sec \ref{architecture}) further improves and enriches the textual descriptions by converting them into textual queries. At the same time, it transforms the ground-truth output into text-based answers, following specific output requirements. By combining these visual graphs, textual queries, and text-based answers, we obtain the Graph-based Vision-Language Question Answering (GVLQA) dataset. 

In the process of establishing GVLQA, we employed graphviz~\citep{graphviz} to instantiate the graph visualizer. This choice is made due to its multitude of pre-defined layout algorithms, which enable convenient adjustment of visual graph layouts. Additionally, manual template-based constructed queries are utilized as the questioner because these tasks are famous with well-defined requirements.

\subsection{Structure}
\label{subsets}
The GVLQA dataset comprises 526K samples, each consisting of a visual graph, a textual query, and its corresponding answer. It is divided into five subsets: GVLQA-BASE, and four augmentation subsets GVLQA-AUGLY,  GVLQA-AUGNS, GVLQA-AUGNO, and GVLQA-AUGET.
In GVLQA-BASE, the visual graphs are uniformly styled. The remaining four augmentation subsets are derived from GVLQA-BASE through the four visual graph augmentations (Sec \ref{visual aug}), varying in six different layouts, three node shapes, four node outline styles, and four degrees of edge thickness, respectively. Detailed statistics of the four subsets are shown in Table \ref{statistic} of Appendix \ref{appendix:statistic}. 

Each GVLQA subset undergoes evaluation across seven graph reasoning tasks, outlined as follows.
\begin{itemize}
\item \textbf{Connectivity}~\citep{sedgewick2001algorithms} (denoted Connect): Determine whether two randomly selected nodes $u$ and $v$ in an undirected graph are connected. 


\item \textbf{Cycle}~\citep{sedgewick2001algorithms}: Identify whether a cycle exists in an undirected graph.


\item \textbf{Topological Sort}~\citep{kahn1962topological} (denoted TS): 
Find a valid topological sort for a directed acyclic graph. Here, topological sort outputs a linear ordering of the nodes such that for every directed edge $u \leftarrow v$, node $u$ comes before $v$ in the ordering.


\item \textbf{Shortest Path}~\citep{dijkstra2022note} (denoted SP): Find the shortest path between two nodes in a weighted undirected graph. The shortest path between two nodes is the path connecting the two nodes with the minimum sum of edge weights along the path.


\item \textbf{Maximum Flow}~\citep{ford1956maximal} (denoted MaxFlow): 
Calculate the maximum flow from a source node to a sink node in a network graph.


\item \textbf{Bipartite Graph Matching}~\citep{karp1990optimal} (denoted BGM): 
Find a matching set in a bipartite graph with the largest number of edges. A matching set is a collection of edges in which no two edges share any common node.


\item \textbf{Hamilton Path}~\citep{gould2003advances} (denoted HP): Find a valid Hamilton path in an undirected graph. A Hamiltonian path is a path that traverses each node in a graph exactly once.

\end{itemize}
Figure \ref{tasks} offers illustrations for these tasks in the GVLQA-BASE dataset. Illustrations of all the GVLQA subsets are provided in Appendix \ref{appendix: GVLQA_Illustration}.

\section{Experiments}
\label{sec:experiments}
\begin{table*}[b]
\vskip -.1in
\centering
\caption{Accuracy (\%) comparisons on GVLQA-BASE under zero-shot and fine-tuning settings, where ``VO'' denotes a variant of GITA using only the vision modality.}
\label{table: main-table}
\resizebox{0.9\textwidth}{!}{
    \begin{tabular}{lccccccccc}
    \toprule
    Models & Connect & Cycle & TS & SP & MaxFlow & BGM & HP & \textbf{Avg} \\
    \midrule
    \multicolumn{9}{c}{\textit{Zero-shot}} \\
    \midrule
    LLaMA2-7B    & \textbf{50.06} & 49.43 & 0.00 & 0.00 & 0.00 & 0.00 & 0.00   & 14.21 \\ 
    Vicuna-7B    & \textbf{50.06} & 49.43 & 0.00 & 0.00 & 0.00 & 0.00 & 0.00   & 14.21 \\ 
    GITA-7B (VO) & \textbf{50.06} & \textbf{50.33} & 0.00 & 0.00 & 0.00 & 0.00 & 0.00 & \textbf{14.34} \\ 
    GITA-7B      & \textbf{50.06} & 49.43 & 0.00 & 0.00 & 0.00 & 0.00  & 0.00  & 14.21 \\ 
    \midrule
    GPT-4 Turbo  & 76.70 & 49.51 & 19.59 & 35.35 & \textbf{6.89} & 42.11 & 47.04 & 39.60\\
    GITA-ZS (VO) & 57.76 & \textbf{63.34} & 5.34 & 4.88 & 1.59 & 46.60 & 10.74 & 27.18\\ 
    GITA-ZS      & \textbf{82.58} & 51.46 & \textbf{19.71} & \textbf{37.69} & 6.00 & \textbf{52.21} & \textbf{50.00} & \textbf{42.81}\\ 
    \midrule
    \multicolumn{9}{c}{\textit{Fine-tuning}} \\
    \midrule
    LLaMA2-7B   & 97.33 & 94.63 & 33.26 & 26.01 & 9.56 & 90.86 & 23.95 &  53.66\\ 
    Vicuna-7B & 97.58 & 95.04 & 34.46 & 25.98 & 9.33 & 91.04 & 25.55   & 54.15 \\ 
    GITA-7B (VO) & 59.97 & 96.34 & 13.30 & 5.72 & 2.89 & 93.01 & 1.11   & 38.91 \\ 
    GITA-7B & \textbf{98.95} & \textbf{96.67} & \textbf{41.12} & \textbf{32.15} & \textbf{20.00} & \textbf{93.19} & \textbf{29.26}   & \textbf{58.76}\\ 
    \midrule
    LLaMA2-13B & 98.79 & 93.36 & 33.83 & 27.93 & 12.22 & 91.34 & 33.46 &  55.85\\ 
    Vicuna-13B & \textbf{99.35} & 94.39 & 36.73 & 28.53 &  11.34 & 92.65 & \textbf{34.81}   & 56.83 \\ 
    GITA-13B (VO)   & 58.00 & \textbf{96.91} & 14.45 & 5.72 & 4.89 & \textbf{93.19} & 1.85   & 39.29 \\ 
    GITA-13B & 99.14 & 95.60 & \textbf{38.69} & \textbf{40.47} & \textbf{20.66} & 92.12 & 33.33   & \textbf{60.00}\\
    \bottomrule
    \end{tabular}
}
\end{table*}

In this section, we extensively evaluate the performance of LLM baselines and the proposed GITA on the GVLQA-BASE and five real-world datasets. To better clarify the reasoning capabilities of solely visual graphs, we also test GITA without the textual descriptions of graphs, which can be considered as a variant of GITA and denoted as vision-only (VO). In this case, the visual graph is the only information source for graph reasoning. Additionally, we investigate the importance of visual graph augmentation (Sec \ref{visual aug}) strategies, by comparison GITA-7B trained on GVLQA-BASE and on the other augmentation subsets of GVLQA (Sec \ref{subsets}). Lastly, we investigate the effectiveness of using GVLQA as the pretrained dataset on real-world datasets. The evaluation metrics for all experiments are accuracy by exact matching. For the fine-tuning setting, we fine-tune the LoRA adapters \citep{lora} for all weight matrices in the text decoder of the VLM reasoner, while keeping the vision encoder in the VLM reasoner frozen. More detailed experimental settings are in Appendix \ref{appendix: experiment setting}.

\subsection{Evaluation on the GVLQA-BASE Dataset}
In this subsection, we perform experiments on the GVLQA-BASE dataset to compare GITA with popular LLMs including GPT-4 Turbo~\citep{gpt4}, LLaMA2-7B/13B~\citep{llama}, and Vicuna-7B/13B~\citep{vicuna}, under both zero-shot and fine-tuning settings. The experimental results are shown in Table \ref{table: main-table}. Based on these results, we can obtain the following observations.

\noindent{\bf Observation 1: GITA Outperforms LLM Baselines.}
As can be seen in Table \ref{table: main-table}, GITA consistently outperforms the LLM baselines under the same setting. This underscores its SOTA effectiveness in instruction-based graph reasoning tasks, showing robust capabilities across different parameter scales under both fine-tuning and zero-shot settings. Moreover, under the fine-tuning setting, incorporating the vision modality consistently benefits 7B models. But for the 13B models, the performance of some tasks may degrade. This could be attributed to the greater challenge of aligning representations of the visual and textual modalities in the larger 13B models compared to the 7B models, in the case of only fine-tuning LoRA adapters in the text decoder. We speculate that full training could potentially address this issue. However, we leave this as future work due to resource constraints.

\noindent{\bf Observation 2: Mainstream Open-source VLM/LLMs Lack Fundamental Graph Reasoning Abilities.}  The zero-shot results illustrate that prominent open-source LLMs or VLMs, including LLaMA2, Vicuna, and LLaVA, exhibit minimal graph reasoning capabilities on the GVLQA-BASE dataset. Specifically, these models produce random answers, i.e., randomly responding with either "Yes." or "No." for tasks involving Connect and Cycle, resulting in a performance close to 50\%. Current SOTA closed-source LLMs or VLMs, including GPT-4 Turbo and GPT-4V, demonstrate superior zero-shot performance compared with the aforementioned open-source models. This observation implies that current open-source LLMs and VLMs lack basic graph reasoning ability, which may be attributed to the insufficient availability of relevant training data. Such observation also enhances our motivation to propose the GVLQA dataset, with the aim of improving the graph reasoning capabilities of VLMs/LLMs.

\noindent{\bf Observation 3: Increasing Model Size Leads to Better Graph Reasoning Capabilities.}
The comparison of VLMs/LLMs with different parameter sizes, specifically 7B  and 13B  models, verify the benefits of increasing the model size for graph reasoning capabilities. In this regard, GITA-13B outperforms its counterpart with 7B parameters (GITA-7B) both on average and across four of the seven tasks. However, it is worth noting that GITA-13B does not outperform GITA-7B on the other three tasks. We hypothesize that this discrepancy may be attributed to insufficient modality alignment due to LoRA fine-tuning.

\noindent{\bf Observation 4: Vision and Text Modalities Proficient in Different Types of Graph Reasoning Tasks.} We explore the individual capabilities of the visual and textual modalities within the GITA framework. The results indicate that the text and vision modalities can complement each other and contribute to better performance than individual ones, as removing either modality leads to performance drops in most cases (Vicuna \& GITA (VO) and GPT-4 Turbo \& GITA (VO) in Table \ref{table: main-table}).
While the graph reasoning capability provided by the vision modality may not be as strong as that of the text modality in most cases, relying solely on vision still enables the model to possess basic graph reasoning abilities. Specifically, the model outperforms text-based LLMs in 2 of the 7 tasks (Cycle and BGM) when relying solely on vision. This consistent improvement across all comparison groups demonstrates the potential of the vision modality to excel in certain graph reasoning tasks, leveraging its ability to capture visual patterns like cycles and graph properties such as bipartition. 
In contrast, text exhibits a higher proficiency than vision modality in sequence-related graph reasoning problems, particularly on tasks such as TS, SP, and HP, which require constructing ordered node sequences.

\subsection{Evaluation for the Visual Graph Augmentations}
\label{augmentation}

\begin{table}[t]
\centering
\caption{Accuracy (\%) comparisons across GVLQA subsets using GITA-7B (VO). $\uparrow$ denotes dramatic performance improvement.}
\label{table: aug_performance}
    \resizebox{0.9\textwidth}{!}{
    \begin{tabular}{lccccccccc}
    \toprule
     & Connect & Cycle & TS & SP & MaxFlow & BGM & HP & \textbf{Avg} \\ 
    \midrule
    GVLQA-BASE     & 59.97 & 96.34 & 13.30 & 5.72 & 2.89 & 93.01 & 1.11    & 38.91 \\
    GVLQA-AUGNS    & 59.85 & 96.75 & 14.17 & 6.61 & 3.78 & 91.58 & 1.48 & 39.17 \\
    GVLQA-AUGNO    & 54.87 & 96.50 & 14.29 & 5.54 & \textbf{3.94} & 92.83 & 1.11 & 38.44 \\
    GVLQA-AUGET    & 57.98 & 96.91 & 13.37 & 5.97 & 3.11 & 91.76 & 0.74 & 38.55 \\
    GVLQA-AUGLY    & \textbf{87.18} $\uparrow$ & \textbf{97.07} & \textbf{14.86} & \textbf{76.55} $\uparrow$ & \textbf{3.94} & \textbf{93.19} & \textbf{70.74} $\uparrow$ & \textbf{63.36} $\uparrow$ \\
    \bottomrule
    \end{tabular}
}
\vskip -.2in
\end{table}

\begin{table}[t]
\centering
\caption{Accuracy (\%) comparisons on real-world datasets under zero-shot and fine-tuning settings, where $\ddagger$ indicates the usage of a checkpoint pretrained in the Cycle task of GVLQA-BASE.}
\label{table: real-world performance}
\resizebox{0.9\textwidth}{!}{
    \begin{tabular}{lcccccc}
    \toprule
    Models & ca-GrQc & ca-HepTh & PolBlogs & Cora & CiteSeer & \textbf{Avg}\\
    \midrule
    \multicolumn{7}{c}{\textit{Zero-shot}} \\
    \midrule 
    LLaMA2-7B & 40.59 & 48.89 & 10.74 & 24.35 & 30.33 & 30.98 \\  
    Vicuna-7B & 41.35 & 50.00 & 8.72  & 26.94 & 29.13 & 31.22 \\
    GITA-7B   & 71.95 & 86.06 & 46.98 & 31.37 & 30.63 & 53.40 \\
    $\text{GITA-7B}^\ddagger$ & \textbf{72.02}   & \textbf{86.08}  & \textbf{48.32} & \textbf{32.10} & \textbf{31.83} & \textbf{54.07} \\
    \midrule 
    \multicolumn{7}{c}{\textit{Fine-tuning}} \\
    \midrule 
    LLaMA2-7B & 76.57 & 89.06 & 80.54 & 83.76 & 73.27 & 80.64\\  
    Vicuna-7B & 78.95 & 89.85 & 80.54 & 84.87 & 74.17 & 81.68\\
    GITA-7B   & 79.70 & 91.13 & 84.56 & 85.24 & 75.07 & 83.14\\
    GITA-7B (w/ AUGLY)  & 79.77  & 91.21 & \textbf{85.23} & 85.24 & 75.68 & 83.43\\
    $\text{GITA-7B}^\ddagger$ & \textbf{80.46} & \textbf{91.68} & \textbf{85.23} & \textbf{86.35} & \textbf{76.57} & \textbf{84.06}  \\ 
    \bottomrule
    \end{tabular}
}
\vskip -.2in
\end{table}
To assess the impact of the proposed visual graph augmentation strategies (including layout, node shape, node outline style, and edge thickness augmentations), we compare the performance of vision-only GITA-7B models trained on the four augmented subsets of GVLQA and on GVLQA-BASE (without augmentation). The results are presented in Table \ref{table: aug_performance}. To fully utilize the visual information in visual graphs, we fine-tune the visual encoder of VLMs in addition to the vision-to-text projector and the LoRA adapters within the text decoder in this experiment.

As can be seen from the results, a significant enhancement in overall performance is observed with the introduction of layout augmentation (GVLQA-AUGLY). The average performance improves remarkably from 38.91\% to 63.36\%. Notably, significant improvements are observed on SP (5.72\% to 76.55\%), HP (1.11\% to 70.74\%), and Connect (59.97\% to 87.18\%). These findings highlight the critical role of layout augmentation in generating visual graphs. In other words, this observation suggests the potential for creating larger-scale datasets for vision-language-based graph reasoning, which could significantly contribute to advancing this field. Conversely, the other three augmentations do not yield such substantial performance improvements, further emphasizing the importance of layout augmentation in vision-language-based reasoning. 

\subsection{Evaluation on Real-World Datasets}
\label{real-world}

In this section, we study the effectiveness of GITA on the ca-GrQC \cite{cadataset} and ca-HepTh \cite{cadataset} datasets for the link prediction task, and on the PolBlog \cite{adamic2005political}, Cora \cite{yang2016revisiting} and CiteSeer \cite{yang2016revisiting} datasets for the node classification task. 
Table \ref{table:real-world statistics} in the appendix \ref{appendix:statistic} presents the statistics of these datasets. The graph can have thousands of nodes/edges, making it infeasible to feed the entire graph into the model. Consequently, we employ \(k\)-hop subgraph sampling 
(with \(k=2\)) 
discussed in Sec \ref{architecture}
to satisfy the token length restriction of LLMs and visual graph scope effectively. 

The experimental results are presented in Table \ref{table: real-world performance}. It is evident that GITA consistently outperforms the LLM baselines, and its performance progressively improves with the addition of layout augmentation and the use of the GVLQA checkpoint. Notably, we emphasize the advantages of using GVLQA-BASE as a pretrained dataset by comparing it with GITA-7b. Performance improvements of 0.67\% and 0.92\% are observed in the zero-shot and fine-tuning settings, respectively. This highlights the potential application value of the proposed GVLQA dataset.

\subsection{Comparison of GITA with Dedicated Graph Baselines}
\begin{table}[!htbp]
\vspace{-20pt}
\centering
\caption{Accuracy (\%) comparisons among dedicated GNNs and GITAs on GVLQA-Base.}
\label{table:gnn}
    \resizebox{0.8\textwidth}{!}{
    \begin{tabular}{lccccccccc}
    \toprule
     & Connect & Cycle & TS & SP & MaxFlow & BGM & HP & \textbf{Avg} \\ 
    \midrule
    GCN  & 79.65 &70.89 & 45.71 &44.56 &\textbf{56.44} &76.70 &32.22 &58.02 \\
       SAGE  & 82.72 &73.58 & 44.51 &\textbf{49.25} &50.67 &81.00 & \textbf{36.67} & 59.78  \\ 
       GITA-7B & 98.95 & \textbf{96.67} & 41.12 & 32.15 & 20.00 & \textbf{93.19} & 29.26 & 58.76 \\
        GITA-13B & \textbf{99.14} & 95.60 & 38.69 & 40.47 & 20.66 & 92.12 & 33.33 & \textbf{60.00} \\
    \bottomrule
    \end{tabular}
}
\vskip -.10in
\end{table}

Though GITA is designed for language-based general graph reasoning settings, which are much more user-friendly (by user-readable natural language) and general (unique model architecture for various scenarios) than the typical application of dedicated GNNs, it remains essential to conduct a comprehensive comparison with specialized GNNs to elucidate the strengths and limitations of GITA's applicability and capabilities. To this end, we assess the graph reasoning abilities of GITA against dedicated GNNs, including GCN \citep{gcn} and SAGE \citep{sage}, using the GVLQA-Base dataset, as detailed in Table \ref{table:gnn}. 
In addition, we explore and compare the effects of \(k\)-hop subgraph sampling on the proposed GITA and GNN baselines. Using the ca-Hepth dataset, we analyze the impact of increasing the number of hops \(k\) on the reasoning time and performance of both GITA and GNNs. The results are in Table \ref{table:gnn2}.

\textbf{\noindent{Overall Graph Reasoning Ability Comparison.}} As shown in Table \ref{table:gnn}, compared to the dedicated GNNs, the fine-tuned GITA-7B models have comparable average graph reasoning performance, with the larger GATA-13B model performs slightly better. In particular, compared to GNNs, the GITA model shows a stronger ability in recognizing local structures in the graphs (Connect and Cycle) and to accomplish tasks with obvious layout heuristics (BGM). We believe that this advantage comes from GITA's visual perception. For SP and MaxFlow, GITA's performance is inferior to GNNs. This may be because GNNs process edge weight more effectively through the message-passing mechanism.

\textbf{\noindent{Scalability and Performance Variation with Different Numbers of Hops \(k\).}}
The inference time results are shown in Table \ref{table:gnn2}. As can be seen, GITA demonstrates inferior scalability compared to the GNN baselines. Its scalability remains stable as the sampled graph size (i.e., \(k\)) increases. 
\begin{wraptable}{r}{0.55\textwidth}
\centering
\caption{Accuracy (\%) and Inference Time (in parentheses) for GNNs and GITA on ca-Hepth Dataset with different subgraph sampling hop number $k\in\{1,2,3,4\}$.}
\label{table:gnn2}
\vskip -.1in
\resizebox{0.55\textwidth}{!}{
\begin{tabular}{cccc}
\toprule
 & \textbf{GCN} & \textbf{SAGE} & \textbf{GITA-7B} \\
\midrule
k=1 & 93.27 (0.02s) & 94.40 (0.03s) & 90.33 (17.23min) \\
k=2 & 94.49 (0.04s) & 94.43 (0.04s) & 91.13 (17.66min)\\
k=3 & 91.10 (0.04s) & 90.95 (0.18s) & 90.31 (17.22min)\\
k=4 & 81.92 (0.05s) & 83.60 (0.22s) & 86.10 (17.01min)\\
\bottomrule
\end{tabular}}
\end{wraptable}
From the accuracy results in Table \ref{table:gnn2}, GITA, GCN, and SAGE achieve their peak performance at \(k = 2\), suggesting that a small sampled graph size suffices for optimal performance. Though the dedicated GNNs attain higher peak performance than GITA, they exhibit performance declines as \(k\) increases (e.g., 3 or 4), while GITA's performance is more stable w.r.t. \(k\).

\subsection{Case Study}
\label{case study}

\begin{wrapfigure}{r}{0.5\textwidth}
\vskip -0.15in
  \centering
  \begin{subfigure}{0.24\textwidth}
    \includegraphics[width=\linewidth]{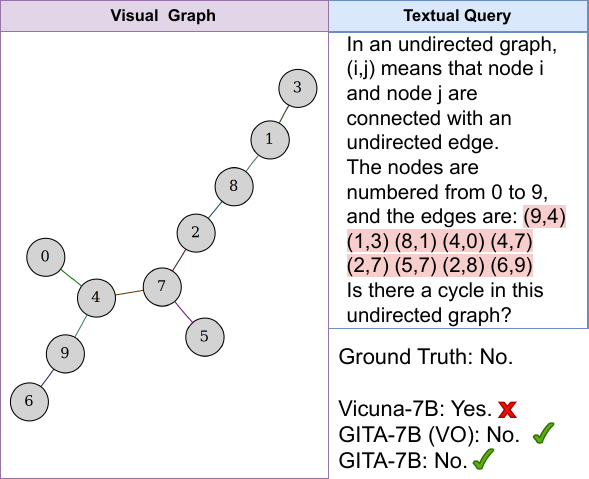} 
    \caption{} 
    \label{fig:sub1}
  \end{subfigure}
  \hfill 
  \begin{subfigure}{0.24\textwidth}
    \includegraphics[width=\linewidth]{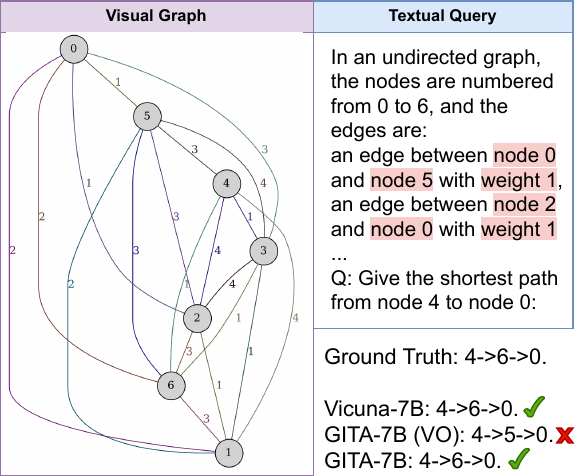} 
    \caption{} 
    \label{fig:sub2}
  \end{subfigure}
  \caption{A comparative case study of graph representation in vision and text modalities. All methods are trained on the GVLQA-BASE dataset.}
  \label{case}
\vskip -.2in
\end{wrapfigure}

In this section, we present examples of graph information provided in both visual and textual formats, which offer some intuitive interpretations for our experimental results. 
Figure \ref{case} (a)
shows an example where the GITA-7B (VO) model outperforms its LLMs-based counterpart,
and 
Figure \ref{case} (b)
shows an opposite scenario.

The task depicted in Figure \ref{case}(a) is cycle detection and the correct answer is `No'. This is predicted successfully by the vision-only GITA-7B model, while the text-based Vicuna-7B fails. In this example, recognizing cycle patterns is much easier in visual graphs, whereas text-based LLMs struggle with disordered textual descriptions of edges, which could inherently involve greater complexity and more challenges.

On the other hand, the fixed layout of visual graphs presented in GVLQA-BASE may impede the visual encoder in identifying the shortest path between two nodes, although we have verified layout augmentation can greatly improve the graph reasoning abilities of models, as shown in Sec. \ref{augmentation}. This limitation might arise from the confusion caused by the visual distance within an image, without considering the weights between the nodes. For instance, in Figure \ref{case}(b), the correct answer is '4->6->0', which visually appears as a more convoluted path but numerically has a shorter path length of \(3=1+2\). In contrast, the incorrect answer given by GITA-7B (vision-only) is '4->2->0', which has a higher path length cost of \(4=1+3\) but visually seems like a more direct shortcut. This observation further validates the effectiveness of employing layout augmentation to enhance performance in this task. Layout variations of visual graphs play a crucial role in mitigating the visual confusion caused by the spatial arrangement within a visual graph. However, it seems more effective for text-based LLMs to handle explicitly separated nodes and weights, as illustrated by the  text 
(in 
red)
in Figure \ref{case}(b).

\section{Conclusion}
In this paper, we propose an end-to-end framework called GITA for vision-language-based graph reasoning. Extensive experiments validate the superiority of incorporating visual information into instruction-based graph reasoning. Furthermore, we conduct comparative analysis of the four proposed visual graph augmentations and identify layout augmentation as the most effective approach for enhancing visual graphs. This finding offers valuable insights for the development of larger-scale datasets aimed at facilitating vision-language-based graph reasoning. Lastly, we highlight the potential application value of the proposed GVLQA dataset as a pretrained dataset.

\section*{Acknowledgements}
This work was supported by NSFC key grant 62136005 and NSFC general grant 62076118, and
in part by the
Research Grants Council of the Hong Kong Special Administrative Region (Grants 16200021 and
16202523).
\bibliography{paper_arxiv}
\bibliographystyle{plain}

\newpage

\appendix
\section{Visual Modality Enhances Effectiveness by Uncovering Critical Substructures}

In this section, we present a case study to highlight the complementary role of the visual modality in graph reasoning tasks. The visual modality excels at recognizing beneficial substructures or local patterns, which are crucial for effective graph reasoning. For instance, identifying the "hop number" is essential for shortest path calculations, recognizing "leaf nodes" is vital for topological sorting, and detecting "cycles" is necessary to avoid in Hamilton path construction. We extracted these substructures from the GVLQA-Base dataset and manually labeled them. By employing a frozen Vision Transformer (ViT) in the LLaVA framework with a trainable Multi-Layer Perceptron (MLP) decoder, we achieved identification accuracies of 89.92\%, 95.16\%, and 92.39\% for hop number counting, leaf node identification, and cycle detection, respectively. In contrast, using a pre-trained BERT model with the same trainable MLP decoder resulted in significantly lower accuracies of 55.47\%, 26.33\%, and 60.32\% for the same tasks. Therefore, the enhanced effectiveness of integrating visual and textual modalities can be attributed to the additional structural information provided by the visual modality, which facilitates the identification of these critical substructures.

\section{Advantages of GITA Over Traditional Graph Neural Networks}
\label{appendix:characteristics}

GITA offers several advantages over traditional Graph Neural Networks (GNNs) in terms of generalizability, flexibility, and user-friendliness:

Unlike GNNs, which require task-specific feature engineering and architecture adjustments, GITA employs a unified model architecture for all tasks, demonstrating its \textbf{generalizability}. By separating task specifications from graph structures, GITA can handle various graph reasoning tasks seamlessly. Additionally, it exhibits strong zero-shot capabilities, allowing it to perform well on tasks it has not been explicitly trained on, which is a feature not commonly found in traditional GNNs.

Besides, traditional GNNs often demand specialized knowledge in model architectures and coding to accommodate diverse tasks, posing a challenge for non-experts. In contrast, GITA overcomes this barrier by employing language-based templates for task adaptation, enhancing its \textbf{flexibility}. This flexibility enables GITA to effectively handle a broad spectrum of tasks, offering a framework that can be customized to specific requirements using daily language, without the necessity of profound expertise in graph neural networks.

Moreover, by leveraging existing VLMs, GITA can respond in natural language, allowing for intuitive graph reasoning with simple queries like "Is there a cycle in this graph?" This stands in contrast to the unreadable vector representations typically used in GNNs, significantly enhancing GITA's \textbf{user-friendliness}.



\section{Datasets Statistics}
\label{appendix:statistic}
\begin{table*}[!htbp]
\centering
\caption{Statistics of the GVLQA dataset.}
\label{statistic}
\resizebox{0.9\textwidth}{!}{%
    \begin{tabular}{lcccccccc}
    \toprule
    Subset & Connect & Cycle & TS & SP & MaxFlow & BMG & HP & Total\\
    \midrule
    BASE & 16,410 &  4,100 &  2,910 &  1,560 &  1,500 & 1,860 & 900 & 29,240\\
    AUGLY & 98,460 & 24,600 & 17,460 & 9,360 & 9,000 & 11,160 & 5,400 & 175,440\\
    AUGNS & 49,230 & 12,300 & 8,730  & 4,680 & 4,500 & 5,580 & 2,700 & 87,720\\
    AUGNO & 65,640 & 16,400 & 11,640 & 6,240 & 6,000 & 7,440 & 3,600  & 116,960\\
    AUGET & 65,640 & 16,400 & 11,640 & 6,240 & 6,000 & 7,440 & 3,600  & 116,960\\ \midrule
    Total & 295,380 & 73,800 & 52,380 & 28,080 & 27,000 &  33,480 & 16,200 & 526,320 \\
    \bottomrule
    \end{tabular}
}
\end{table*}

\begin{table*}[!htbp]
\centering
\caption{Average numbers of nodes and edges for each task in GVLQA.}
\begin{tabular}{lccccccc}
\toprule
Average / Task & Connect & Cycle & TS & SP & MaxFlow & BGM & HP \\ 
\midrule
\#nodes & 25.01 & 23.42 & 21.86 & 13.65 & 13.90 & 21.13 & 13.24 \\ 
\#edges & 95.46 & 23.66 & 114.10 & 23.99 & 49.16 & 51.03 & 45.05 \\ \bottomrule
\end{tabular}

\label{table:avg_nodes_edges}
\end{table*}

\begin{table*}[!hbtp]
\centering
\caption{Statistics of real-world datasets}
\label{table:real-world statistics}
\resizebox{0.8\textwidth}{!}{%
    \begin{tabular}{cccccc}
    \toprule
     & ca-GrQC & ca-HepTh & PolBlogs & Cora & CiteSeer\\ 
     \midrule
    \# Nodes & 5,242 & 9,877 & 1,490 & 2,708 & 3,327\\
    \# Edges & 14,496 & 25,998 & 19,025 & 5,278 & 4,676\\
    domain  & collaboration  & collaboration & social & citation  & citation\\
    average degree & 5.53 & 5.26 & 25.54 & 3.9  & 2.74\\
    
    \bottomrule
    \end{tabular}%
}
\end{table*}

\section{Illustrations for Visualization Tools in Graph Visualizer}
\label{appendix: visualizer implement}

The GITA graph visualizer incorporates a variety of implementations for existing visualization tools such as Graphviz, Matplotlib with NetworkX, and igraph, each selected for their unique capabilities in graph rendering. These tools are implemented in our code as interchangeable modules, enhancing flexibility based on the requirements of different projects. 

\begin{figure}[!htbp]
  \centering
  \includegraphics[width=0.9\textwidth]{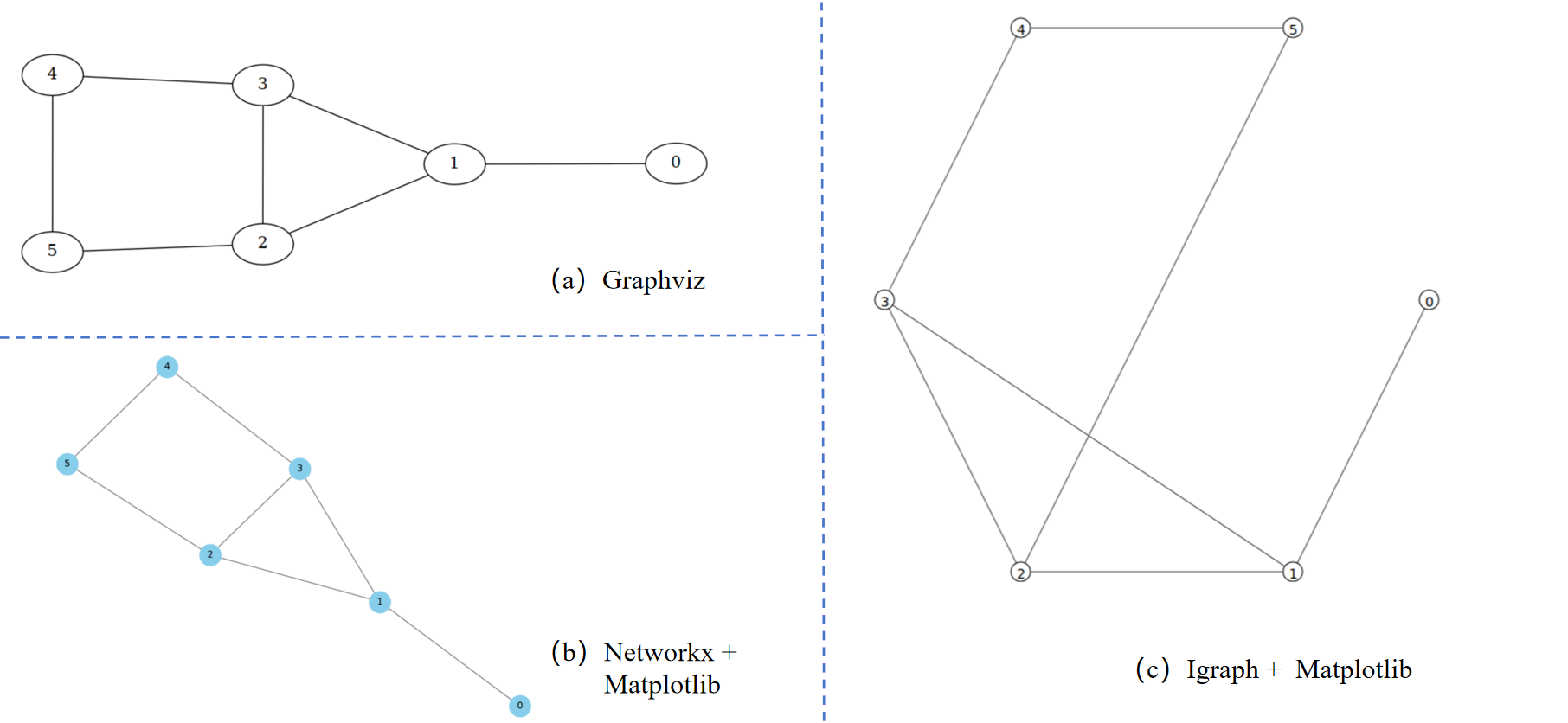}
        \vskip -.1in
    \caption{Examples of the visual graph generated by various visualization tools.} 
    \label{fig:diff_visualizer}
\end{figure}
Figure \ref{fig:diff_visualizer} showcases some visual graphs produced by these different graph visualizer implementations.

\section{Graph-describing Templates}
\label{appendix: describer template}
The graph describer relies on a set of unified structured templates designed to generate coherent and informative descriptions that emphasize the inherent characteristics of the graph itself, regardless of specific task requirements. These graph-describing templates cover various scenarios, including directed graphs, undirected graphs, graphs with node identities or features, and graphs with edge weights or capacities. Table \ref{graph templates} provides an illustration of these templates, where [P] denotes placeholders required to be filled by corresponding graph information.

\begin{table*}[!htbp]
    \footnotesize 
    \centering 
    \setlength\tabcolsep{1.5pt}
    \renewcommand{\arraystretch}{2}
\resizebox{\linewidth}{!}{
    \begin{tabular}{p{1in} p{3in} p{3in}} 
    \toprule 
    \textbf{Graph categories} &  \textbf{Undirected} & \textbf{Directed}   \\
    \midrule
    Prototype & In an undirected graph, (i,j) means that node i and node j are connected with an undirected edge. The nodes are numbered from [P] to [P], and the edges are: 
    \newline  
    ([P], [P]) , ([P], [P])... & 
    In a directed graph, (i,j) means that node i and node j are connected with a directed edge from node i to node j. The nodes are numbered from [P] to [P], and the edges are: 
    \newline         ([P], [P]) , ([P], [P])... \\ 
    \midrule
    W/ Node Attributes  & In an undirected graph, the nodes are numbered from [P] to [P], and every node has an attribute. (i,j) means that node i and node j are connected with an undirected edge.
    \newline The attributes of nodes are:
    \newline node [P]: [P]
    \newline node [P]: [P]
    \newline ...
    \newline The edges are: ([P],[P]) ([P],[P]) ... &  In a directed graph, the nodes are numbered from [P] to [P], and every node has an attribute. (i,j) means that node i and node j are connected with a directed edge from node i to node j.
    \newline The attributes of nodes are:
    \newline node [P]: [P]
    \newline node [P]: [P]
    \newline ... 
    \newline The edges are: ([P],[P]) ([P],[P]) ...   \\
    \midrule
    W/ Edge Weights & In an undirected graph, the nodes are numbered from [P] to [P], and the edges are:\newline an edge between node [P] and node [P] with weight [P],\newline an edge between node [P] and node [P] with weight [P], 
    \newline ... & In a directed graph, the nodes are numbered from [P] to [P], and the edges are:\newline an edge from node [P] to node [P] with weight [P],\newline an edge from node [P] to node [P] with weight [P], 
    \newline ...\\ 
    \midrule
    W/ Both & In an undirected graph, the nodes are numbered from [P] to [P], and every node has an attribute. 
    \newline The attributes of nodes are: 
    \newline node [P]: [P]
    \newline node [P]: [P]
    \newline ... 
    \newline And the edges are:
    \newline an edge between node [P] and node [P] with weight [P],
    \newline an edge between node [P] and node [P] with weight [P], 
    \newline ... & In a directed graph, the nodes are numbered from [P] to [P], and every node has an attribute. 
    \newline The attributes of nodes are: 
    \newline node [P]: [P]
    \newline node [P]: [P]
    \newline ... 
    \newline And the edges are:
    \newline an edge from node [P] to node [P] with weight [P],
    \newline an edge from node [P] to node [P] with weight [P], 
    \newline ... \\ 
    \bottomrule
\end{tabular}}
\caption{Graph-describing Templates for various categories.}
\label{graph templates}
\end{table*}

\section{Examples of Manual-template-based and LLM-agent-bootstrapped Query Generation}
\label{appendix: Questioner}
\begin{figure}[!htbp]
  \centering
  \includegraphics[width=\textwidth]{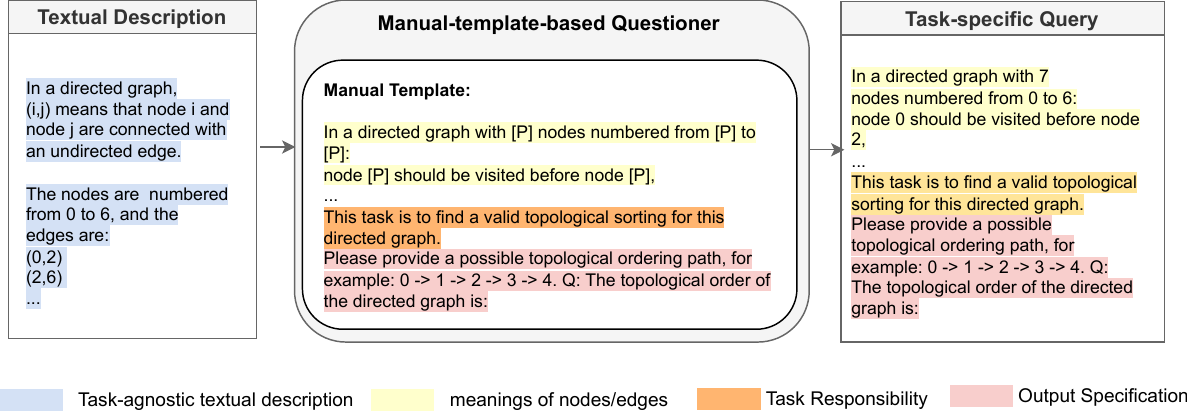}
        \vskip -.1in
    \caption{Examples of the manual-template-based query generation, where [P] denotes the placeholders.} 
    \label{fig:manual_gene}
\end{figure}

\noindent{\bf Manual-template-based Query Generation.} The queries $Q^T_G$ can be generated by task-specific manual templates. These templates are manually crafted by human to supplement descriptions/instructions about 1) concrete meanings of nodes and edges, 2) task
responsibilities and 3) input/output specifications into the task-agnostic graph description $D_G$. Therefore, the \textbf{precision} and \textbf{faith} of generated task-specific queries $Q^T_G$ are guaranteed by human calibrations. An example of manual-template-based query generation for topological sorting is illustrated in Figure \ref{fig:manual_gene}. In this example, placeholders [P] are used to represent information that scripts will automatically fill in. 

\begin{figure}[!htbp]
  \centering
  \includegraphics[width=\textwidth]{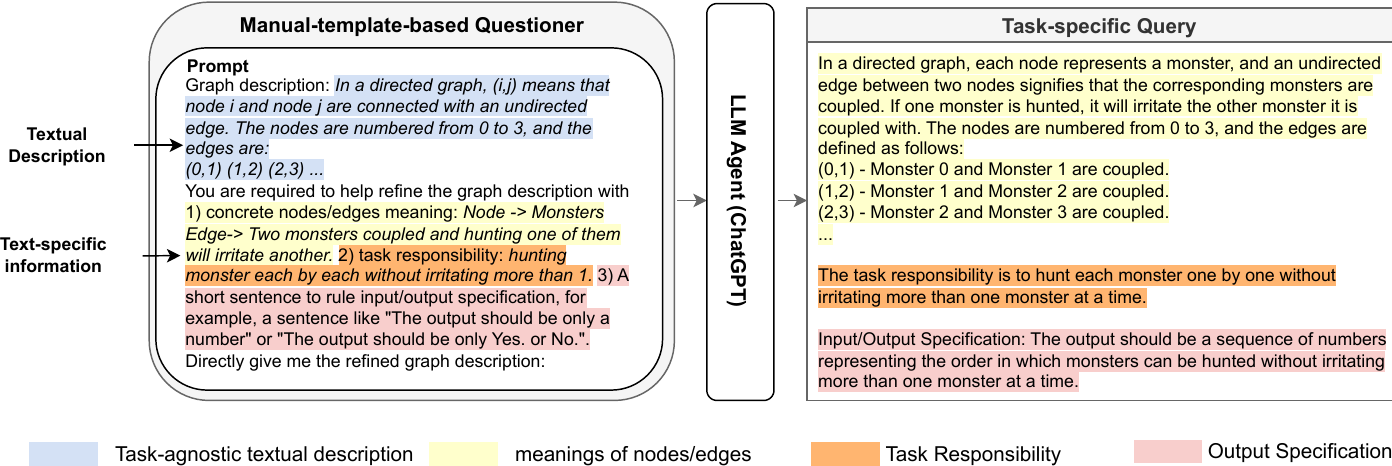}
        \vskip -.1in
    \caption{Examples of the LLM-agent-bootstrapped query generation.} 
    \label{fig:llm_gene}
\end{figure}

\noindent{\bf LLM-agent-bootstrapped Query Generation.}  Figure \ref{fig:llm_gene} presents an example of employing a bootstrapped LLM agent, such as ChatGPT\citep{chatgpt}, for monster-hunting gaming. By incorporating task-specific information into the prompt, including node/edge meanings and task responsibilities, the LLM agent automatically generates a response that serves as the desired task-specific query. Compared to using manual templates, bootstrapping LLM agents for automated synthesis is more \textbf{flexible} and \textbf{economic} as it can take advantage of LLM agents to automatically interpret context and generate appropriate queries for various scenarios and minimize manual effort with changing conditions. Such properties make it suitable for dynamic or bespoke tasks, such as robotic planning or complex gaming scenarios.

\section{Experiment Settings}
\label{appendix: experiment setting}
For all fine-tuning experiments, we use a batch size of 128 and adopt the AdamW optimizer (with a learning rate of 0.0002 and 0.00002 for the LoRA adapters within the text decoder and vision-to-text projector, respectively). 

\noindent{\bf Detailed Settings for GVLQA Dataset}
During the evaluation, the temperature is set to 0 for all baselines. All fine-tuning experiments are conducted on an NVIDIA DGX station with 8×A100 GPUs. We split the GVLQA dataset in the ratio of 7:3 for training and testing, respectively. The accuracy (\%) metrics are computed by comparing the prediction and ground truths with exact matching. We use the next-token-prediction loss to fine-tune the LoRA \citep{lora} adapters of LLMs and the vision-to-text projector. Visual graphs are encoded as visual embeddings by a visual encoder. Visual embeddings are concatenated with the embeddings of textual descriptions and instructions
(i.e., questions), then fed to the text decoder to generate the answer.

\noindent{\bf Real-world Datasets}
Here we provide more details about the five real-world datasets used in Sec \ref{real-world}. The datasets ca-GrQC and ca-HepTh represent collaboration networks from the arXiv sections of General Relativity and Quantum Cosmology, and High Energy Physics - Theory, respectively, featuring nodes as authors and edges as co-authorships. They can be downloaded from Stanford Network Analysis Project (SNAP) website \footnote{https://snap.stanford.edu/index.html}. PolBlogs is a network of U.S. political blogs from February 2005, categorized by political alignment and linked by blog references. Cora and CiteSeer are both citation networks, where nodes correspond to scientific papers and edges to citations, utilized for tasks such as document classification and citation analysis, with papers categorized into various research fields. 
Statistics of the datasets are shown in Table \ref{table:real-world statistics}.
For each dataset,
$80\%/10\%/10\%$ of the edges 
are randomly 
used
for 
training/validation/testing, 
respectively.

\noindent{\bf Detailed Settings for Real-world Benchmarks} In the conventional semi-supervised node classification setting, class labels are available for some nodes, which is reflected in the visual graph by coloring the nodes with a unique random color for each class. To focus on evaluating the model's ability to capture structural information, GITA filters out the influence of node features in Cora and CiteSeer datasets.
For link prediction tasks on ca-GrQC and ca-HepTh datasets, GITA treats the graphs as undirected. In the test split, both the original links and their reverse links do not appear in the train and valid splits. During training and evaluation, an equal number of negative sampled links are used alongside the positive links. These negative links are sampled at each training epoch but remain fixed during evaluation.
For the GVLQA pretrain checkpoint, GITA adopts the 7B cycle checkpoint finetuned on GVLQA-BASE, where the performance is nearly mature. Hyperparameter combinations for each model are determined through grid search, and the specific combinations can be found in the provided code.

\section{Illustrations of GVLQA subsets}
\label{appendix: GVLQA_Illustration}
\begin{figure*}[!htbp]
\centering
\includegraphics[width=0.98\textwidth]{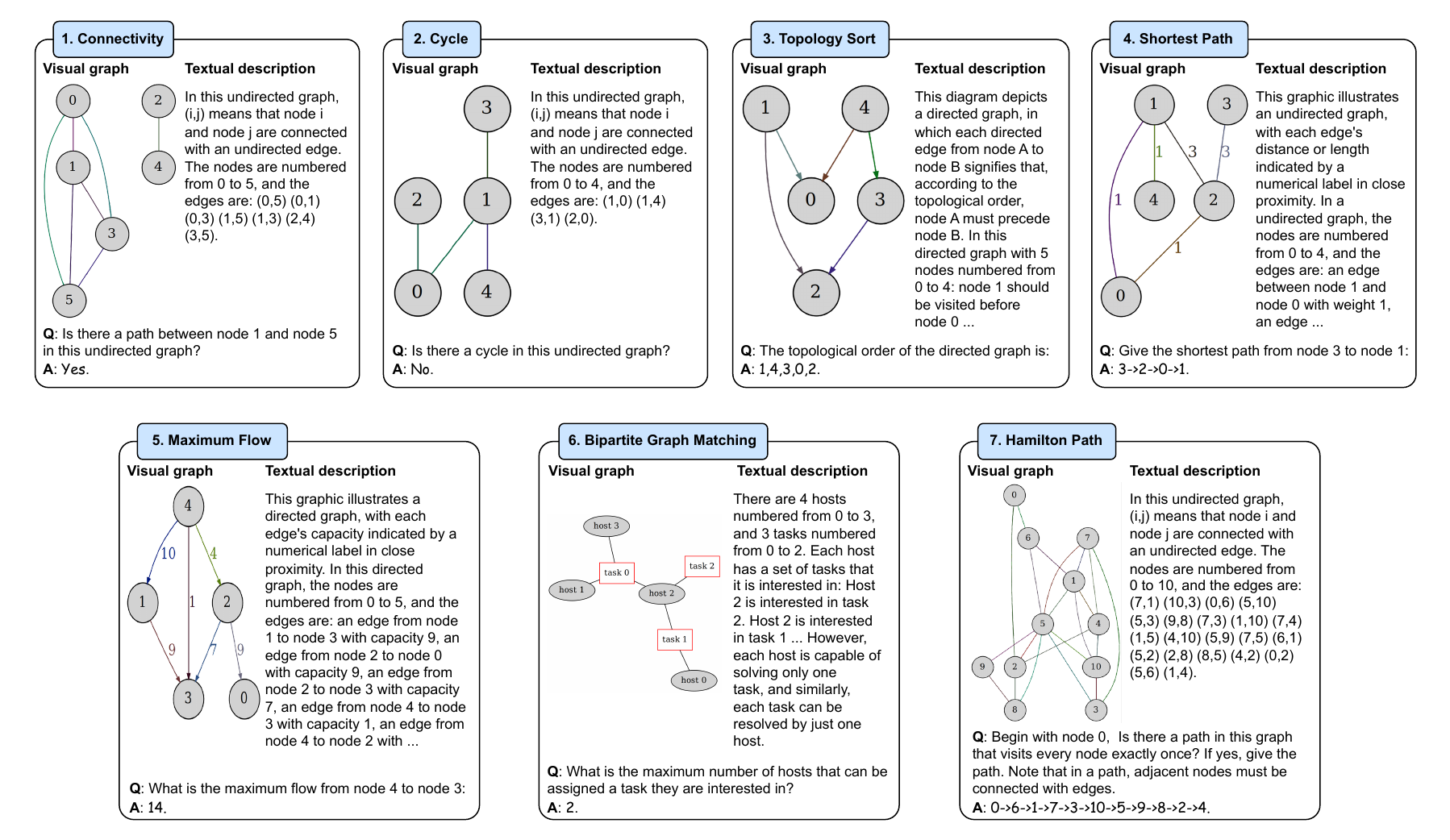}
\caption{An overview of the GVLQA-BASE. Each figure depicts the tasks involving graph-based reasoning, showcasing a visual graph, a textual question, and the corresponding answer. }
\label{tasks}
\end{figure*}

\begin{figure*}[!htbp]
\centering
\includegraphics[width=0.98\textwidth]{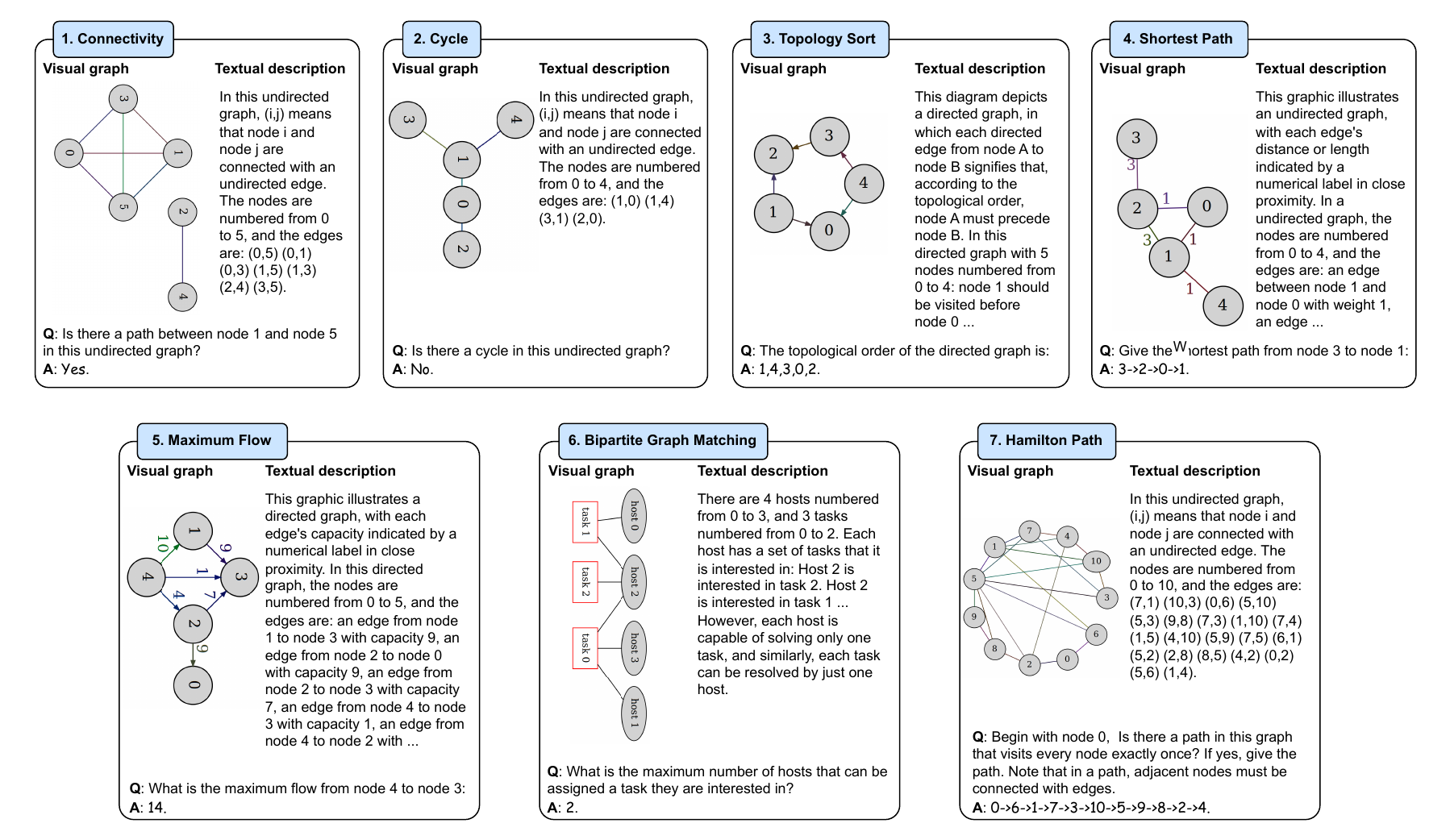}
\caption{An overview of the GVLQA-AUGLY. Figures are akin to GVLQA-BASE but vary only in layouts. }
\label{AUGLY}
\end{figure*}

\begin{figure*}[!htbp]
\centering
\includegraphics[width=0.98\textwidth]{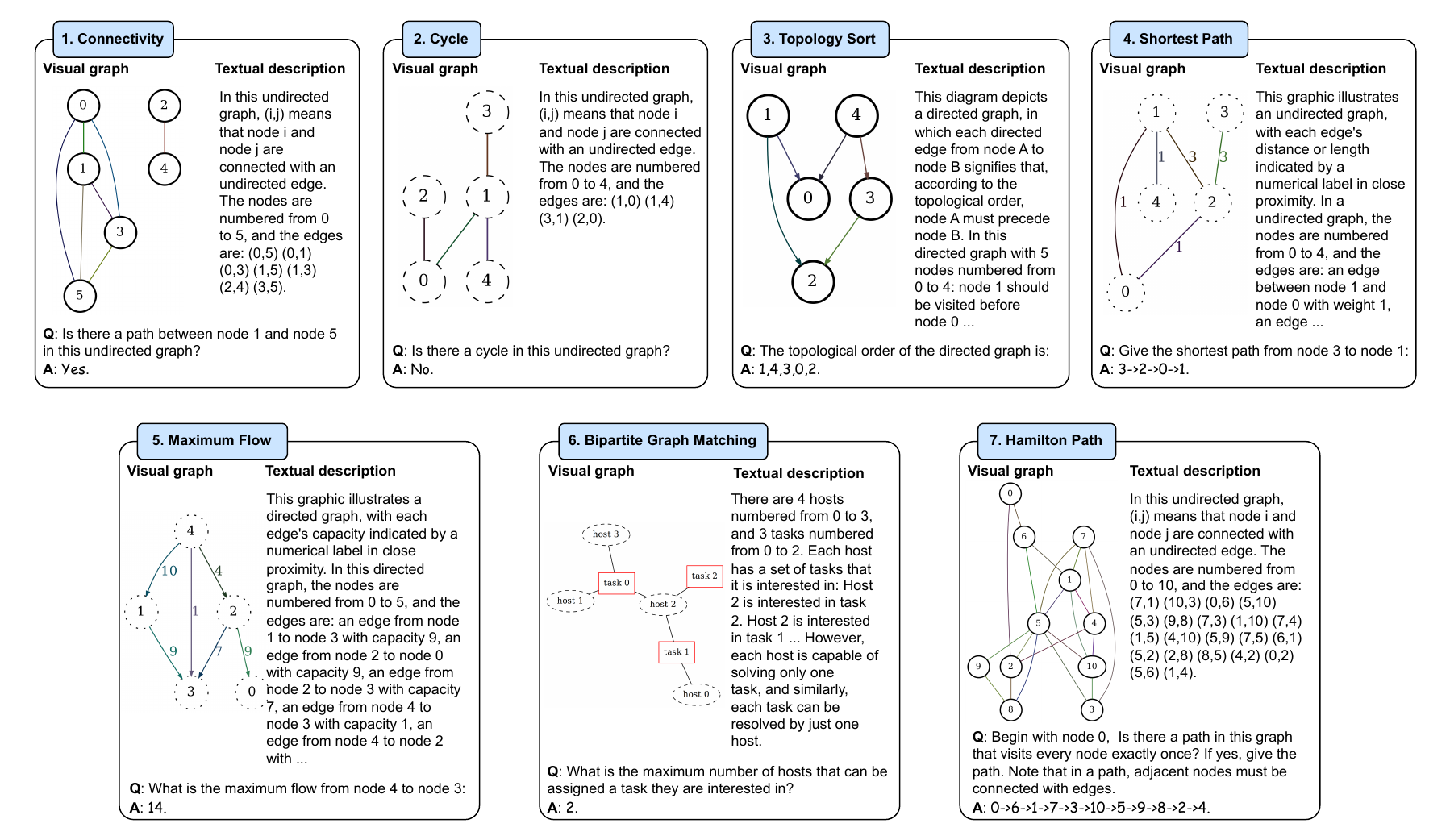}
\caption{An overview of the GVLQA-AUGNO. Figures are akin to GVLQA-BASE but vary only in node outline styles. }
\label{AUGNO}
\end{figure*}

\begin{figure*}[!htbp]
\centering
\includegraphics[width=0.98\textwidth]{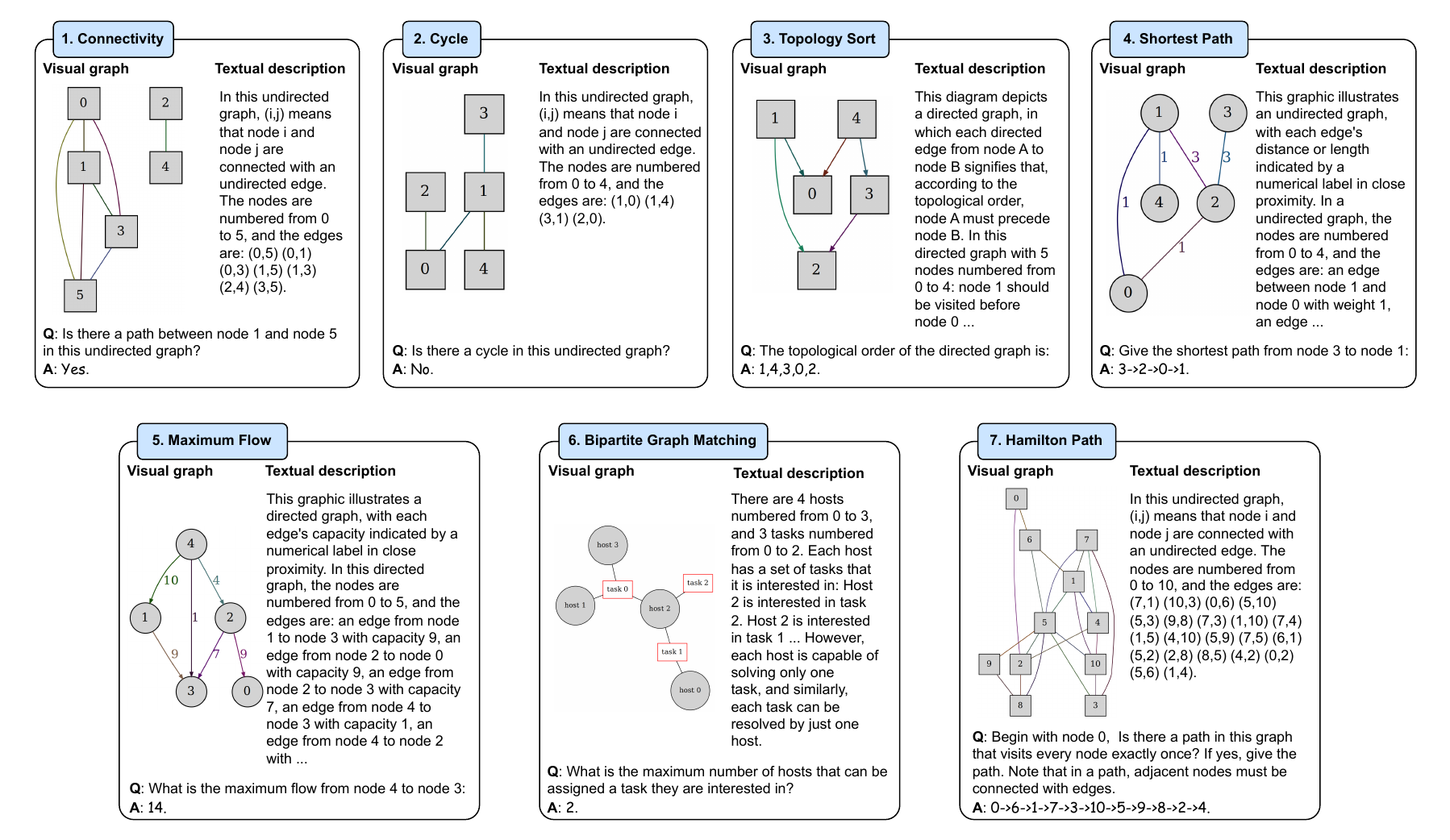}
\caption{An overview of the GVLQA-AUGNS. Figures are akin to GVLQA-BASE but vary only in node shapes. }
\label{AUGNS}
\end{figure*}

\begin{figure*}[!t]
\centering
\includegraphics[width=0.98\textwidth]{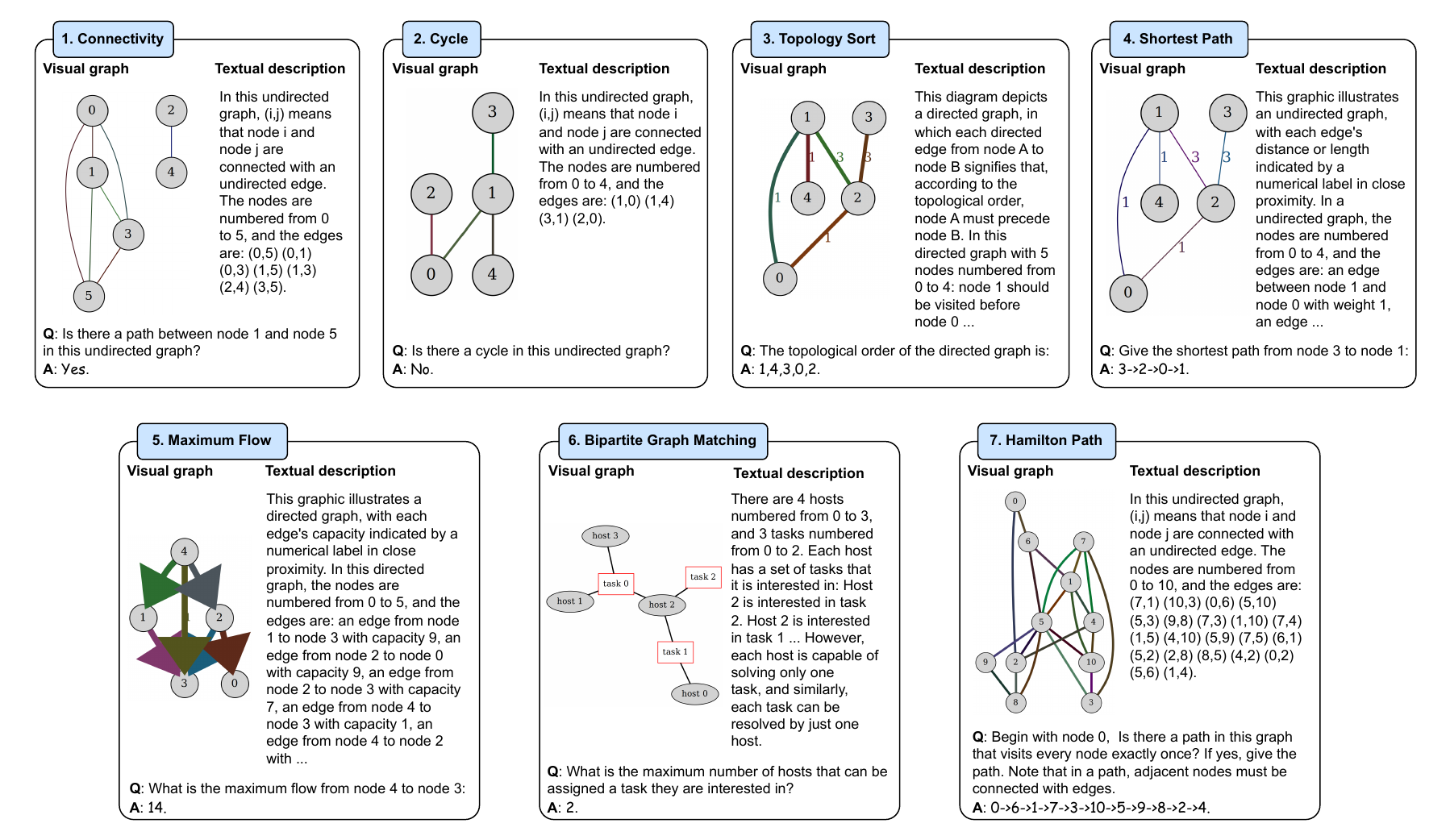}
\caption{An overview of the GVLQA-AUGET. Figures are akin to GVLQA-BASE but vary only in edge thicknesses. }
\label{AUGET}
\end{figure*}

In this section, we present the illustrations of the GVLQA subsets. Figure \ref{tasks} provides an overview of GVLQA-BASE. Subsequently, from Figure \ref{AUGLY} to Figure \ref{AUGET}, we showcase the augmented visual graphs in GVLQA-AUGLY (augment layouts), GVLQA-AUGNS (augment node styles), GVLQA-AUGNO (augment node outline styles), and GVLQA-AUGET (augment edge thickness), respectively.

\section{Limitation}
\label{appendix:limitation}
The GITA framework proposed in the paper, along with its experimental results, exhibits certain limitations that should be acknowledged. Firstly, when dealing with large-scale graphs, the conventional subgraph sampling strategy employed by GITA may result in imbalanced and insufficient sampling, leading to the loss of critical graph structural information. This compromise is necessary to accommodate the limited contextual length of the text-based LLM and the restricted scope of the visual graph. Secondly, due to computational constraints, the fine-tuning procedures in the paper were restricted to the LoRA framework. While this approach offers advantages, a more comprehensive fine-tuning process that considers both visual and text modalities is expected to better align the two and potentially enhance performance. Addressing these limitations should be considered as a future research direction in this field.

\clearpage

\section*{NeurIPS Paper Checklist}

The checklist is designed to encourage best practices for responsible machine learning research, addressing issues of reproducibility, transparency, research ethics, and societal impact. Do not remove the checklist: {\bf The papers not including the checklist will be desk rejected.} The checklist should follow the references and follow the (optional) supplemental material.  The checklist does NOT count towards the page
limit. 

Please read the checklist guidelines carefully for information on how to answer these questions. For each question in the checklist:
\begin{itemize}
    \item You should answer \answerYes{}, \answerNo{}, or \answerNA{}.
    \item \answerNA{} means either that the question is Not Applicable for that particular paper or the relevant information is Not Available.
    \item Please provide a short (1–2 sentence) justification right after your answer (even for NA). 
\end{itemize}

{\bf The checklist answers are an integral part of your paper submission.} They are visible to the reviewers, area chairs, senior area chairs, and ethics reviewers. You will be asked to also include it (after eventual revisions) with the final version of your paper, and its final version will be published with the paper.

The reviewers of your paper will be asked to use the checklist as one of the factors in their evaluation. While "\answerYes{}" is generally preferable to "\answerNo{}", it is perfectly acceptable to answer "\answerNo{}" provided a proper justification is given (e.g., "error bars are not reported because it would be too computationally expensive" or "we were unable to find the license for the dataset we used"). In general, answering "\answerNo{}" or "\answerNA{}" is not grounds for rejection. While the questions are phrased in a binary way, we acknowledge that the true answer is often more nuanced, so please just use your best judgment and write a justification to elaborate. All supporting evidence can appear either in the main paper or the supplemental material, provided in appendix. If you answer \answerYes{} to a question, in the justification please point to the section(s) where related material for the question can be found.

IMPORTANT, please:
\begin{itemize}
    \item {\bf Delete this instruction block, but keep the section heading ``NeurIPS paper checklist"},
    \item  {\bf Keep the checklist subsection headings, questions/answers and guidelines below.}
    \item {\bf Do not modify the questions and only use the provided macros for your answers}.
\end{itemize}


\begin{enumerate}

\item {\bf Claims}
    \item[] Question: Do the main claims made in the abstract and introduction accurately reflect the paper's contributions and scope?
    \item[] Answer: \answerYes{} 
    \item[] Justification: Please refer to the abstract part and the contribution enumeration at the tail of the introduction part.

\item {\bf Limitations}
    \item[] Question: Does the paper discuss the limitations of the work performed by the authors?
    \item[] Answer: \answerYes{} 
    \item[] Justification: Please refer to Appendix \ref{appendix:limitation}

\item {\bf Theory Assumptions and Proofs}
    \item[] Question: For each theoretical result, does the paper provide the full set of assumptions and a complete (and correct) proof?
    \item[] Answer: \answerNA{} 
    \item[] Justification: The paper does not involve any theoretical result.

    \item {\bf Experimental Result Reproducibility}
    \item[] Question: Does the paper fully disclose all the information needed to reproduce the main experimental results of the paper to the extent that it affects the main claims and/or conclusions of the paper (regardless of whether the code and data are provided or not)?
    \item[] Answer: \answerYes{} 
    \item[] Justification: We show fundamental experiment settings in Section \ref{sec:experiments}, and more details for experiments settings in Appendix \ref{appendix: experiment setting}. Besides, we provide the complete codes as supplementary materials.

\item {\bf Open access to data and code}
    \item[] Question: Does the paper provide open access to the data and code, with sufficient instructions to faithfully reproduce the main experimental results, as described in supplemental material?
    \item[] Answer: \answerYes{} 
    \item[] Justification: The complete codes are included, and the proposed GVLQA dataset is released with common access.
    \item[] Guidelines:
    \begin{itemize}
        \item The answer NA means that the paper does not include experiments requiring code.
        \item Please see the NeurIPS code and data submission guidelines (\url{https://nips.cc/public/guides/CodeSubmissionPolicy}) for more details.
        \item While we encourage the release of code and data, we understand that this might not be possible, so “No” is an acceptable answer. Papers cannot be rejected simply for not including code, unless this is central to the contribution (e.g., for a new open-source benchmark).
        \item The instructions should contain the exact command and environment needed to run to reproduce the results. See the NeurIPS code and data submission guidelines (\url{https://nips.cc/public/guides/CodeSubmissionPolicy}) for more details.
        \item The authors should provide instructions on data access and preparation, including how to access the raw data, preprocessed data, intermediate data, and generated data, etc.
        \item The authors should provide scripts to reproduce all experimental results for the new proposed method and baselines. If only a subset of experiments are reproducible, they should state which ones are omitted from the script and why.
        \item At submission time, to preserve anonymity, the authors should release anonymized versions (if applicable).
        \item Providing as much information as possible in supplemental material (appended to the paper) is recommended, but including URLs to data and code is permitted.
    \end{itemize}

\item {\bf Experimental Setting/Details}
    \item[] Question: Does the paper specify all the training and test details (e.g., data splits, hyperparameters, how they were chosen, type of optimizer, etc.) necessary to understand the results?
    \item[] Answer: \answerYes{} 
    \item[] Justification: Experiment details are given in both Section \ref{sec:experiments} and Appendix \ref{appendix: experiment setting}.

\item {\bf Experiment Statistical Significance}
    \item[] Question: Does the paper report error bars suitably and correctly defined or other appropriate information about the statistical significance of the experiments?
    \item[] Answer: \answerNo{} 
    \item[] Justification: The paper does not include error bars.

\item {\bf Experiments Compute Resources}
    \item[] Question: For each experiment, does the paper provide sufficient information on the computer resources (type of compute workers, memory, time of execution) needed to reproduce the experiments?
    \item[] Answer: \answerYes{} 
    \item[] Justification: We report the machine (type and storage) requirements in Appendix \ref{appendix: experiment setting}.
    
\item {\bf Code Of Ethics}
    \item[] Question: Does the research conducted in the paper conform, in every respect, with the NeurIPS Code of Ethics \url{https://neurips.cc/public/EthicsGuidelines}?
    \item[] Answer: \answerYes{} 
    \item[] Justification: We make sure the research conducted in the paper conform, in every respect, with the NeurIPS Code of Ethics.

\item {\bf Broader Impacts}
    \item[] Question: Does the paper discuss both potential positive societal impacts and negative societal impacts of the work performed?
    \item[] Answer: \answerNA{}{} 
    \item[] Justification: The research does not have concerns about societal impacts because it is designed for general-purpose graph reasoning.
    
\item {\bf Safeguards}
    \item[] Question: Does the paper describe safeguards that have been put in place for responsible release of data or models that have a high risk for misuse (e.g., pretrained language models, image generators, or scraped datasets)?
    \item[] Answer: \answerYes{} 
    \item[] Justification: The paper includes using an graph visualizer to generate abstract graph images, however, these images are focus on graph structure, without any sensitive information.

\item {\bf Licenses for existing assets}
    \item[] Question: Are the creators or original owners of assets (e.g., code, data, models), used in the paper, properly credited and are the license and terms of use explicitly mentioned and properly respected?
    \item[] Answer: \answerYes{} 
    \item[] Justification: We have cited necessary assets and conduct CC-BY 4.0 for our codes and datasets.

\item {\bf New Assets}
    \item[] Question: Are new assets introduced in the paper well documented and is the documentation provided alongside the assets?
    \item[] Answer: \answerYes{} 
    \item[] Justification: The code and other supplementary materials are followed with readme and instructions.

\item {\bf Crowdsourcing and Research with Human Subjects}
    \item[] Question: For crowdsourcing experiments and research with human subjects, does the paper include the full text of instructions given to participants and screenshots, if applicable, as well as details about compensation (if any)? 
    \item[] Answer: \answerNA{} 
    \item[] Justification: the paper does not involve crowdsourcing nor research with human subjects.

\item {\bf Institutional Review Board (IRB) Approvals or Equivalent for Research with Human Subjects}
    \item[] Question: Does the paper describe potential risks incurred by study participants, whether such risks were disclosed to the subjects, and whether Institutional Review Board (IRB) approvals (or an equivalent approval/review based on the requirements of your country or institution) were obtained?
    \item[] Answer: \answerNA{} 
    \item[] Justification: the paper does not involve crowdsourcing nor research with human subjects.

\end{enumerate}

\end{document}